\documentclass{article}

    \PassOptionsToPackage{numbers, compress}{natbib}

    \usepackage[final]{icbinb2021}

\usepackage[utf8]{inputenc} %
\usepackage[T1]{fontenc}    %
\usepackage{hyperref}       %
\usepackage{url}            %
\usepackage{booktabs}       %
\usepackage{amsfonts}       %
\usepackage{nicefrac}       %
\usepackage{microtype}      %
\usepackage{xcolor}         %

\usepackage{amssymb}
\usepackage{mathtools}
\usepackage[makeroom]{cancel}

\DeclareMathOperator{\E}{\mathbb E}
\DeclareMathOperator*{\argmin}{argmin}
\DeclareMathOperator*{\argmax}{argmax}

\newcommand{\D}{\mathcal D}

\newcommand{\entropy}{\mathbb H}
\newcommand{\obj}{\mathcal L}

\newcommand{\KL}[2]{D_\text{KL}\left(#1 \mid \mid #2 \right)}

\newcommand{\var}{\mathrm{Var}}

\newcommand{\prob}{\mathbb P}

\title{Entropic Issues in Likelihood-Based OOD Detection}

\author{%
  Anthony L. Caterini \\
  Layer 6 AI \\
  \texttt{anthony@layer6.ai} \\
   \And
   Gabriel Loaiza-Ganem \\
   Layer 6 AI \\
   \texttt{gabriel@layer6.ai} \\
}

\begin{document}

\maketitle

\begin{abstract}
  Deep generative models trained by maximum likelihood remain very popular methods for reasoning about data probabilistically.
  However, it has been observed that they can assign higher likelihoods to out-of-distribution (OOD) data than in-distribution data, thus calling into question the meaning of these likelihood values.
  In this work we provide a novel perspective on this phenomenon, decomposing the average likelihood into a KL divergence term and an entropy term.
  We argue that the latter can explain the curious OOD behaviour mentioned above, suppressing likelihood values on datasets with higher entropy.
  Although our idea is simple, we have not seen it explored yet in the literature.
  This analysis provides further explanation for the success of OOD detection methods based on likelihood ratios, as the problematic entropy term cancels out in expectation.
  Finally, we discuss how this observation relates to recent success in OOD detection with manifold-supported models, for which the above decomposition does not hold directly.
\end{abstract}

\section{Introduction}

Deep Generative Models (DGMs) are a staple for probabilistic modelling of high-dimensional data, and are typically trained by maximum likelihood -- or approximations thereof -- over observations from some ground truth distribution.
One possible application of DGMs is identifying whether new data is in- or out-of-distribution (OOD), which na\"ively seems quite straightforward given a sufficiently powerful model: average likelihood could always be further increased by raising it for observed points, and thus lowering it for OOD data.
However, \citet{nalisnick2019deep} provided a landmark study showing that DGMs do not behave in this way, as models trained on the CIFAR-10 dataset \citep{krizhevsky2009learning} produce higher likelihoods when tested on Street View House Numbers (SVHN) \citep{netzer2011reading}, and similarly for models trained on Fashion-MNIST \citep{xiao2017fashion} and tested on MNIST \citep{lecun1998gradient}.
In either case, we note that the phenomenon only occurs in one direction, with the more complicated dataset of the pair (i.e.\ CIFAR-10 or Fashion-MNIST) consistently scoring lower likelihoods than the simpler one (i.e.\ SVHN or MNIST).
Various explanations have been posed for this phenomenon, including OOD data residing in high likelihood but low probability regions \citep{nalisnick2019deep, nalisnick2019detecting}, early layers of the architectures encoding generic information \citep{kirichenko2020normalizing, schirrmeister2020understanding}, poor model fit \citep{zhang2021understanding}, or representations of the data being being inadequate \citep{just2019deep, le2020perfect}.

In this paper, we propose a simple, alternative explanation for this phenomenon, centred on a standard decomposition of the average likelihood into (i) a KL divergence term between the true data distribution and the model, and (ii) an entropy term.
The latter is constant with respect to the parameters of the DGM, and is thus easily ignored in optimization, but can wildly vary over different datasets.
In particular, data distributions with high entropy may be biased towards lower likelihoods, providing a new perspective on how even a perfect DGM could be unable to perform OOD detection.
This interpretation further explains the success of OOD detection methods based on likelihood ratios, which we see are invariant to the entropy term in expectation.
We also examine how our analysis relates with the typical set approach \citep{nalisnick2019detecting}, which itself only considers the entropy of the in-distribution set and not the incoming test data, thus becoming exposed to pathological cases where likelihoods are similar but the data is far different.
As an addendum to the main analysis of this work, we note that the likelihood decomposition does not apply as-is to DGMs supported on low-dimensional manifolds embedded in the higher-dimensional ambient space.
We conjecture that this may account for some of the recent improvements in OOD detection for models of this type, as they may be less susceptible to the influence of the entropy term.

\section{Background}

\paragraph{Deep Generative Models} DGMs are, loosely speaking, models which combine probabilistic elements with deep neural networks.
In this paper, we focus on DGMs which admit densities\footnote{Either with respect to the Lebesgue measure for continuous data, or the counting measure for discrete data.} and are trained by maximum likelihood -- or an approximation thereof -- such as normalizing flows \citep{papamakarios2019normalizing, kobyzev2020normalizing}, variational auto-encoders (VAEs) \citep{DBLP:journals/corr/KingmaW13, rezende2014stochastic}, and energy-based models \citep{lecun2006tutorial, du2019implicit}.
DGMs are often used in practice as \emph{density estimators}, since, in expectation, maximizing a model's likelihood over an observed dataset also minimizes its KL divergence from the true distribution generating this data (cf.\ \eqref{eq:main_decomposition}).
Likelihood-based DGMs have demonstrated practical success across a wide variety of domains (e.g.\ \citet{kingma2014semi, gao2016linear, oord2018parallel, townsend2018practical, NEURIPS2020_e3b21256}), which might suggest that the likelihood values outputted by these models are reliable proxies for the probability that new data is similar to the data on which the model was trained.

\paragraph{Failures on Out-of-Distribution Detection}

However, \citet{nalisnick2019deep} discovered that this is \emph{not} the case, as DGMs can often assign higher likelihoods to data which is dissimilar to the training set, or out-of-distribution (OOD), particularly if this new data is considered to be ``simpler'' than the training set.
For example, as mentioned earlier, models trained on CIFAR-10 produce higher likelihoods when evaluated on SVHN data, with similar phenomena observed for other pairs of datasets.
This observation has led researchers to question the meaning of likelihood values outputted by DGMs in an attempt to both understand why this phenomenon occurs and what (if anything) can be done about it; we discuss approaches along either of these directions throughout this work.

\section{The Influence of Entropy on the Likelihood}

In this section, we provide a simple and straightforward explanation for this strange OOD phenomenon based on the standard decomposition of the likelihood into a KL divergence term and an entropy term.
We argue that the latter can inflate or deflate likelihoods, thus making these values unreliable on their own.
Throughout this section we assume all distributions admit densities, either with respect to the Lebesgue measure on the ambient space for continuous data, or the counting measure for discrete data; our analysis covers both cases.
This agrees with standard assumptions made when building full-support DGMs, and we will discussion the implications of relaxing this in \autoref{sec:discussion}.

\subsection{Decomposition of the Likelihood} \label{sec:decomposition}

Suppose we have a dataset $\mathcal D_P \coloneqq \{x_i\}_{i=1}^n$, with each $x_i$ sampled i.i.d.\ from an unknown data-generating distribution $P$.
Suppose also that we have a model for this distribution parametrized by $\theta$, denoted $P_\theta$, which admits a density $p_\theta$.
We can write the typical likelihood objective to estimate $\theta$ as $\sum_{i=1}^n \log p_\theta(x_i)/n$, which can be rewritten in the limit of infinite data as
\begin{align} \label{eq:main_decomposition}
    \obj(\theta) \coloneqq \lim_{n\rightarrow \infty} \frac 1 n \sum_{i=1}^n \log p_\theta(x_i) = \E_{X\sim P}[\log p_\theta(X)] = - \KL{P}{P_\theta} - \entropy[P],
\end{align}
where $D_\text{KL}$ denotes the KL divergence and $\entropy$ denotes the entropy.
This well-known result can motivate maximum likelihood estimation -- indeed, \citet{nalisnick2019deep} provide a similar decomposition in (1) of their work -- as we have $\argmax_\theta \obj(\theta) = \argmin_\theta \KL P {P_\theta}$ since $\entropy[P]$ is independent of $\theta$.

However, this result also provides a straightforward explanation for why we see strange behaviour when evaluating the likelihood of a model on a \emph{different} dataset $\mathcal D_Q \coloneqq \{y_j\}_{j=1}^m$, with each $y_j$ sampled i.i.d.\ from an unknown distribution $Q$ with the same support as $P$.
We can again write the average likelihood of $P_\theta$ over $\D_Q$ asymptotically as
\begin{equation} \label{eq:likelihood_ood}
    \lim_{m\rightarrow\infty} \frac 1 m \sum_{j=1}^m \log p_\theta(y_j) = \E_{Y \sim Q} [\log p_\theta(Y)] = -\KL{Q}{P_\theta} - \entropy[Q].
\end{equation}
Now, suppose $Q \neq P$, so that $\D_Q$ is considered OOD.
If $\entropy[P]$ is much larger than $\entropy[Q]$, we can easily have $\E_{Y \sim Q} [\log p_\theta(Y)] > \E_{X\sim P}[\log p_\theta(X)]$ \emph{even if $P_\theta$ is a perfect model of $P$.}
Qualitatively, CIFAR-10 (resp.\ Fashion-MNIST) is far more complex -- and thus ostensibly has higher entropy -- than SVHN (resp.\ MNIST),  which succinctly explains the observation that models trained on the former score higher likelihoods on the latter.
\citet{nalisnick2019deep} additionally note that constant (and thus low-entropy) inputs score very highly on likelihood, agreeing with the analysis above.
At any rate, this discussion provides further insight on the unreliability of using only the likelihood as a test statistic to evaluate whether new data is in- or out-of-distribution.

\subsection{From OOD Detection Failure on Average to Failure with High Probability\protect\footnote{This subsection was added after publication and is not part of the NeurIPS workshop version of this paper.}}

The previous section explains why likelihood-based OOD detection can fail \emph{on average}, i.e.\ that $\E_{Y \sim Q}[\log p_\theta (Y)] > \E_{X \sim P}[\log p_\theta (X)]$ when $\KL{Q}{P_\theta} + \entropy[Q] < \KL{P}{P_\theta} + \entropy[P]$, which as discussed can happen even when $\KL{Q}{P_\theta} > \KL{P}{P_\theta}$, provided $\entropy[Q]$ is much smaller than $\entropy[P]$.
Yet
this does not \emph{immediately} imply that for independent random variables $X \sim P$ and $Y \sim Q$, the inequality $\log p_\theta(Y) > \log p_\theta(X)$ holds with high probability.
However, in this section, we argue that we can indeed obtain a high-probability bound from the result in expectation. %

To simplify notation, let $Z \coloneqq \log p_\theta(Y) - \log p_\theta(X)$ with mean $\mu \coloneqq \KL{P}{P_\theta} + \entropy[P] - \KL{Q}{P_\theta} - \entropy[Q]$, and with variance $\sigma^2 \coloneqq \var(\log p_\theta(X)) + \var(\log p_\theta(Y)) > 0$, which we assume to be finite.
We want to show that $Z > 0$ holds with high probability under suitable assumptions.
Recall Chebyshev's inequality, which states that $\prob (|\mu - Z| \geq k\sigma) \leq 1/k^2$ for any $k > 0$. Assuming that $\mu > 0$, it follows from Chebyshev's inequality that:
\begin{equation}
    \prob (Z > 0) = \prob\left(\mu - Z < \dfrac{\mu}{\sigma} \sigma\right) \geq \prob\left(|\mu - Z| < \dfrac{\mu}{\sigma} \sigma\right) \geq 1 - \dfrac{\sigma^2}{\mu^2}.
\end{equation}
The relationship between $\mu$ and $\sigma$ depends on $P_\theta$, $P$, and $Q$, and it is not necessarily the case that $\sigma^2/\mu^2$ is very close to zero, though here we argue that it is indeed possible.
If $\mu$ is very large (i.e.\ if likelihood-based OOD detection fails on average due to $\entropy[P]$ being much larger than $\entropy[Q]$) and both in-distribution and OOD log-likelihoods do not vary too widely (i.e.\ if $\var(\log p_\theta(X))$ and $\var(\log p_\theta(Y))$ -- and thus $\sigma^2$ -- are relatively small), then likelihood-based OOD detection will fail with high probability.
Our derivation is once again consistent with the empirical observations of \citet{nalisnick2019deep}: likelihoods of models trained on CIFAR-10 (resp. Fashion-MNIST) are often lower -- not just on average -- when evaluated on in-distribution data than on SVHN (resp. MNIST).

\subsection{Likelihood Ratios for OOD Detection Cancel Out the Entropy} \label{sec:likelihood-ratios}

The above analysis suggests that performing likelihood-based OOD detection without somehow accounting for the entropy of incoming data is problematic.
On the other hand, OOD detection methods based on likelihood \emph{ratios} have recently demonstrated strong performance \citep{ren2019likelihood, serra2020input, schirrmeister2020understanding}.
Here, we argue that such approaches intrinsically control for this entropy.
Suppose we have a model $P_\theta$ with density $p_\theta$ trained on data from $P$, and we also have access to some \emph{reference} model $R_\phi$ admitting a density $r_\phi$;
\citet{ren2019likelihood} forms this as a model for the background, \citet{schirrmeister2020understanding} trains this on millions of unrelated images, and \citet{serra2020input} uses standard image compression techniques such as PNG as a proxy for this distribution.
The above methods assess incoming data $y$ as OOD if they score poorly on the likelihood ratio $\log (p_\theta(y) / r_\phi(y))$.
Now, if we assume $y \sim Q$ for some unknown $Q$, we can take the expectation of the likelihood ratio, following \eqref{eq:likelihood_ood}:
\begin{align} \label{eq:likelihood_ratio}
    \E_{Y \sim Q} \log p_\theta(Y) - \E_{Y \sim Q} \log r_\phi(Y) &= - \KL{Q}{P_\theta} - \cancel{\entropy[Q]} + \KL{Q}{R_\phi} + \cancel{\entropy[Q]}.
\end{align}
In particular, we note that this does not explicitly depend on the entropy of incoming data $\entropy[Q]$.
This might explain the success of likelihood ratio methods in OOD detection, as \eqref{eq:likelihood_ratio} will be mediated by the distance from $Q$ to $P$, assuming $P_\theta$ is a good model of $P$ and $\KL{Q}{R_\phi} \lessapprox \KL{P}{R_\phi}$.

\section{Related Work} \label{sec:related}

As far as we are aware, nobody has considered the effect of the entropy term on OOD detection.
We will review some alternative explanations here and provide further discussion.

\paragraph{Perfect Models} \citet{le2020perfect} also noted that OOD detection methods based on likelihood scoring are unreliable \emph{even when provided with a perfect density model of in-distribution data}, although under the observation that likelihood-based OOD detection techniques are not invariant to different representations of the same data.
Our work instead focuses on the impact of the entropy term in OOD detection and thus provides a complimentary analysis; perhaps this suggests that we should look for lower-entropy representations of data before performing  likelihood-based OOD detection.

\paragraph{Likelihood Ratios} \citet{bishop1994novelty} is the earliest instance of a likelihood ratio approach for novelty detection of which we are aware, with much more recent work heading in this direction as mentioned in \autoref{sec:likelihood-ratios}. 
Of note is that \citet{serra2020input} and \citet{schirrmeister2020understanding} require trained models on data of the same generic type but not the same particular distribution -- an image compression algorithm and a density model on 80 million tiny images, respectively -- thus requiring additional knowledge besides just the in-distribution training data.
However, other works \citep{le2020perfect, zhang2021understanding} have indeed suggested that some knowledge of the out-distribution is crucial to attacking the OOD detection problem.
\citet{ren2019likelihood} do not require additional data, but still require training a \emph{background} model on randomly perturbed in-distribution data.
Noting that the above do not work well for VAEs,
\citet{xiao2020likelihood} develop a successful OOD detection method comparing the likelihood of the posterior optimized for a single observation against the likelihood from the entire VAE.
All of these approaches control for the entropy of incoming data as previously discussed in \autoref{sec:likelihood-ratios}, and further can assess \emph{individual} inputs as in- or out-of-distribution which is an attractive property for practitioners.

\paragraph{Typicality}
An alternative explanation of the unintuitive OOD behaviour of likelihood-based models is presented in \citet{nalisnick2019detecting}, who note that a probability distribution's regions of high likelihood may not be associated with regions of high probability -- especially as dimensionality increases.
This led to the development of a test characterizing in- and out-of-distribution on the basis of \emph{typicality}, i.e.\ how similar test (log-)likelihood values are to those from  in-distribution training data.
This is akin to controlling for the entropy of the in-distribution data.
However, as per \autoref{sec:decomposition}, we could imagine a scenario where we observe OOD data with similar likelihoods to the in-distribution data, but which also has far different KL and entropy terms that throw off the comparison.
Controlling for the entropy of the \emph{incoming} data would help prevent such a scenario from occurring.

Another approach entirely is to consider test statistics other than the log likelihood, yet still comparing on the basis of typicality.
\citet{choi2018waic} themselves provide an early discussion on typicality, and find that an ensemble-based estimate of the Watanabe-Akaike Information Criterion \citep{watanabe2010asymptotic} performs well empirically.
\citet{sastry2020detecting} decide whether or not incoming data is OOD based on how its pairwise feature correlations compare with those on in-distribution data.
\citet{morningstar2021density} propose an approach which takes any reasonable statistic(s) from a pre-existing DGM and builds a density estimator on this, rejecting points as OOD if they are not likely under this model.
These methods are fully unsupervised in that they do not require access to OOD data, but often require additional model training or isolating seemingly random features of the data and thus make their general application less clear. 

\paragraph{Model Fit Failure} The final perspective on OOD detection that we will cover centres around model fit failure.
\citet{kirichenko2020normalizing} focus on the inductive biases of normalizing flow methods, remarking that these learn generic image features and thus overpower the maximum likelhood objective as it relates to OOD detection; \citet{schirrmeister2020understanding} find something similar in that the likelihoods on convolutional-based architectures are dominated by low-level features.
These methods propose modified architectures or test statistics, respectively, demonstrating improved results.
\citet{zhang2021understanding} argue that model fit is the most likely culprit in OOD detection failures, and highlights several issues with the (typical set) idea that the support of OOD data could overlap with that of the training data and thus assumes that this does not occur.
This is yet another complementary analysis to ours: in the view of \eqref{eq:main_decomposition} and \eqref{eq:likelihood_ood}, assuming we have a model $P_\theta$ which matches $P$ perfectly, and $P$ and $Q$ have non-overlapping support, then $\KL{Q}{P_\theta} = \infty$ and thus $\E_Q \log p_\theta(Y) \rightarrow -\infty$.
Since we do not observe this in practice, it suggests that model fit may indeed be part of the issue.

\section{Discussion and Conclusion} \label{sec:discussion}

In this work, we have presented an alternative explanation for the strange phenomenon wherein a deep generative model trained on one dataset assigns high likelihood to OOD data, particularly when this data is simpler than the training data.
By exploiting a well-known decomposition of the average likelihood, we find that these observations can be explained by fluctuations in the entropy term between datasets.
This also sheds further light on the success of OOD detection methods based on likelihood \emph{ratios}, as these statistics are not affected by the entropy term (in expectation).

Although we find this observation interesting on its own, one limitation of this work is that we have not devised a new method to overcome issues posed by the entropy term, beyond the suggestion to use pre-existing likelihood ratio approaches.
We at least hope that bringing this explanation to light might inspire other researchers to consider the impact of the entropy term when performing OOD detection with deep generative models.
We would also like to specifically probe this phenomenon in future work with experiments and compare the relative impacts of the KL and entropy terms on the likelihood, but will need to get around the issues with estimating entropy in high dimensions \citep{lombardi2016nonparametric}.

This work may also serve as a motivation for further work in OOD detection with likelihood-based methods supported on low-dimensional manifolds embedded in high-dimensional ambient space, as the decomposition \eqref{eq:main_decomposition} does immediately not apply in this case.
Recently, \citet{caterini2021rectangular} have shown that injective flow methods (e.g.\ \citep{brehmer2020flows}) can accurately assign lower likelihoods to MNIST when trained by maximum likelihood over Fashion-MNIST; perhaps these methods are not as susceptible to fluctuations in the entropy term and can provide further improvements to likelihood-based OOD detection.
Indeed, in accordance with the \emph{manifold hypothesis} \citep{bengio2013representation}, it seems unlikely that high-dimensional data such as images would even admit densities with respect to the Lebesgue measure on the ambient space, and thus low-dimensional density models appear to be an important direction forward.

\begin{ack}
We would like to acknowledge Charline Le Lan, Harry Braviner, and Rob Cornish for their very helpful discussions.
\end{ack}

\bibliographystyle{plainnat}
\bibliography{refs.bib}

\end{document}